\documentclass[conference,a4paper]{IEEEtran}
\IEEEoverridecommandlockouts

\usepackage[hidelinks]{hyperref}
\usepackage[cmex10]{amsmath}
\usepackage{amssymb,amsfonts}
\usepackage{dblfloatfix}

\usepackage[ruled,vlined]{algorithm2e}
\usepackage{graphicx}
\graphicspath{{Figures/PDF/}{Figures/PNG/}}
\usepackage{float}  

\usepackage{booktabs}
\usepackage{siunitx}
\usepackage[numbers,compress]{natbib}
\usepackage{texnames}
\usepackage{bm,bbm}
\usepackage{orcidlink}
\usepackage{multirow}
\begin{document}

\title{\uppercase{Evaluating Graph Neural Networks for Change-Criticality Classification in Maritime Navigation Charts}
}

 \author{
 \textit{
 Abhishek Potnis\textsuperscript{a}\orcidlink{0000-0001-8168-857X}, 
 Jacob Arndt\textsuperscript{a}\orcidlink{0000-0002-1097-0428}
}\\[1ex]
 \textsuperscript{a}\textit{Geospatial Science and Human Security Division, Oak Ridge National Laboratory, USA.}\\
 \{potnisav, arndtjw\}@ornl.gov
 }

\maketitle
\begin{abstract}
Graph neural networks (GNNs) are a class of neural networks suitable for learning on graph-structured data. Their application to spatial data is a natural extension, however its relatively unclear which message-passing operations, architectural configurations, and graph representation is best suited for classifying changes to objects in electronic navigational charts (ENCs)--geospatial vector datasets used for marine navigation. Maintaining these datasets is a challenge, and categorizing changes to objects in the ENC based on their significance to navigational safety is of particular importance. Here, we propose to represent these vector navigation datasets as a graph structure where the spatial objects serve as nodes and their spatial and semantic relationships form edges. We encode both the old ENC dataset and new ENC dataset into a pair of graphs and frame the task as a graph-pair classification problem. Building on this representation, we investigate the use of GNN architectures to classify whether the encoded graphs constitutes a \textit{critical} or \textit{non-critical} risk to navigational safety. We train and evaluate several GNN architectures and model configurations on ENC changes reviewed by maritime experts. Our results demonstrate that graph-based representations improve the classification of ENC updates, providing a scalable approach for automating or improving ENC maintenance workflows.

\end{abstract}


\section{Introduction} 
Electronic Navigational Charts (ENCs) \cite{NOAA_ENC} are geospatial vector datasets used for marine navigation and describe real-world entities relevant for safe navigation at sea. Derived from hydrographic surveys and other geospatial sources, ENCs typically contain  hundreds to thousands of records, each representing an object that is further described by a class category, geometry, and key-value attribute pairs. Table~\ref{tab:example_enc_objects} provides several example objects present in ENCs. One significant challenge is maintaining these charts, which requires determining whether changes to the ENC datasets reflect critical or non-critical risks to navigational safety. In this work, we approach this problem by modeling ENC updates as structured graphs and evaluating graph neural networks for automated critical-change classification, providing a scalable, data-driven solution.

\begin{table}[!h]    \centering
        \caption{Examples of objects in ENCs with some examples of possible attributes.}
    \begin{tabular}{lll}
        \toprule
        \textbf{Class} & \textbf{Geometry} & \textbf{Attributes} \\
        \midrule
         Sounding & Point & Depth \\
         Coastline & Line & Category of coastline, \\
         Light & Point & Category of light, Color, Height \\
         Depth Area & Polygon & Depth value 1, Depth value 2\\
         Qaulity of Data & Polygon & Category of quality \\
         Wreck & Point, Polygon & Category of wreck \\
         \bottomrule
    \end{tabular}
\label{tab:example_enc_objects}
\end{table}

ENC data includes heterogeneous geometries (points, lines, polygons), contains semantic relationships between objects (aggregation peers, association peers, master-slave hierarchies), include spatial topological relationships that need to be considered (depth contours touching depth areas), mixed attribute types (numbers, free form text, coded strings), object-attribute dependencies, and cartographic scale dependencies. Determining whether a given ENC change represents a critical risk to navigational safety is a challenging and time-consuming task, requiring expertise in marine navigation, cartography, and the ENC data product specification. To aid human analysts in their review of ENC changes, we seek to develop a machine learning model to automatically classify these changes. 

We propose representing ENC datasets and their changes as graphs, where ENC objects are encoded as nodes and both spatial and ENC standard relationships are encoded as edges. Following this graph construction, we use graph neural networks (GNN) \cite{scarselli2008graph} to learn to classify differences in the old and new graphs. GNNs are a class of deep learning models designed to operate directly on graph-structured data by iteratively aggregating and transforming information from a node’s neighbors. Leveraging GNNs allows us to capture the rich relational and spatial context inherent in ENC data, which is difficult to model with traditional approaches. In this study, we apply GNNs to classify whether changes in the graph should be considered \textit{critical} or \textit{non-critical}. The goal of this benchmarking study is to evaluate and compare the performance of different GNN architectures on a fixed ENC graph dataset for the task of ENC change-criticality classification, focusing on architectural variations while maintaining a consistent dataset and experimental setup.

The contributions of this paper are: 
\begin{itemize}
    \item We develop a simple and effective graph representation for ENC changes data.
    \item We systematically evaluate multiple GNN architectures along with architecture depth and connectivity to assess their impact on graph-based ENC change-criticality classification performance.
\end{itemize}




\section{Methods}
\subsection{Problem Formulation}
Our objective is to classify ENC changes as \textit{critical} or \textit{non-critical} based on their significance to navigational safety at sea. Given as input, an old ENC dataset, a new ENC dataset, and a change that identifies which objects have been added, deleted, or modified between datasets, we seek to predict whether the change is a \textit{critical} or \textit{non-critical} risk to navigational safety. In the context of graph representation learning, we frame this problem as a graph-pair classification task where the old and new ENC datasets are represented as graph structures to capture the spatial and relational context.

\subsection{Graph Dataset Construction}
Graphs are a natural data structure for representing spatial data and we extend this idea to representing ENC changes as graphs. ENC data has explicit semantic relationships defined by the ENC data product specification that can be modeled as edges in a graph. Furthermore, the spatial context of ENC objects are important and can also be modeled by edges in a graph. To construct an ENC change graph, we take as input the old ENC dataset, new ENC dataset, and identified change, and produce a pair of graphs. Each graph \( G = (V, E, R) \), where \( V \) are nodes of the same type (ENC objects), \( E \) is the set of edges, and \( R \) is the set of edge types which include spatial relationships and  ENC standard relationships.

\textbf{Nodes.} We represent each ENC object as a node in the graph. We intentionally use a minimal node feature design, consisting only of the categorical ENC object class identifier for each node. This choice isolates the contribution of the graph structure itself and avoids confounding effects from richer semantic or geometric attributes.

\textbf{Edges.} We construct edges between nodes by using their S-57 ENC standard relationships and spatial relationships. Four semantic edge types are derived from the S-57 standard: \textit{master–slave}, \textit{slave–peer}, \textit{aggregation–peer}, and \textit{association–peer}. Three spatial relationship edges are also included, based on binary topological predicates: \textit{overlaps}, \textit{touches}, and \textit{equals}. Applying this edge construction to both the old and new ENC datasets produces two corresponding graph structures. Given the limited prior work applying GNNs to ENC data, for this research study, we use a unified edge set that merges all semantic and spatial relationships. This choice simplifies the architecture and allows us to assess the value of graph structure without introducing relation-specific message passing.

\subsection{Architectures}
We consider a simple non-graph baseline and several prominent graph neural network models in our experiments. We describe these below.

\textbf{Baseline.} We develop a simple multilayer perceptron (MLP) baseline model for benchmarking our graph-based approaches. Following \cite{gorishniy2021revisiting}, this MLP baseline consists of 4 linear layers each with dimension 256. The ENC object class and change type (add, delete, change, multifeature change, duplicate feature change) are represented as categorical indices and mapped to learnable dense embeddings via separate embedding layers. This data encoding and model combination does not explicitly incorporate information about the spatial context or ENC relationships for ENC objects in the change.

We evaluate multiple GNN architectures for the task of ENC change classification, comparing their ability to leverage graph structure beyond the baseline MLP. The architectures considered include:

\textbf{Graph Convolutional Network (GCN) \cite{kipf2016semi}.} A spectral-based convolutional GNN that aggregates information from neighboring nodes via normalized adjacency.

\textbf{Graph Attention Network (GAT) \cite{velivckovic2018graph}.}  Introduces attention weights on edges to learn the relative importance of neighboring nodes during message passing.

\textbf{GraphSAGE \cite{hamilton2017inductive}.} An inductive GNN that aggregates neighborhood information using mean or pooling operations, enabling generalization to unseen nodes.


\textbf{Graph Transformer (TransformerConv) \cite{shi2020masked}.} Combines GNN-style message passing with multi-head attention over node neighborhoods. While inspired by transformers, this layer operates on local graph neighborhoods rather than globally attending over all nodes.

\subsection{GNN Architecture Considerations}

To systematically study architectural effects, we vary the number of layers in each GNN and incorporate layer-wise residual connections:

\textbf{Number of Layers}: Varying the number of layers affects the receptive field of nodes and the extent of information propagation across the graph, influencing performance and the risk of over-smoothing.

\textbf{Residual Connections}: Layer-wise residual connections help maintain stable training in deeper GNNs by improving gradient flow and mitigating performance degradation.

All GNN architectures share the same node features, unified edge set, and global features for a controlled comparison, allowing us to isolate the effects of depth and residual connections on predictive performance.








\subsection{Graph-Pair Classification Model for ENC Change Classification}

We implement a Siamese-style \cite{bromley1993signature} graph-pair model for ENC change classification. Each graph, corresponding to the old or new ENC dataset, is passed independently through the same GNN encoder (GCN, GAT, GraphSAGE, or TransformerConv). This produces graph-level embedding that capture local graph structure.
Global ENC change metadata, including the ENC object class for the changed object and change type, are encoded as categorical indices and mapped to learnable embeddings, providing contextual information beyond the node features.

The graph embedding from the old and new graphs are subtracted to form a difference representation that explicitly captures changes. Any mismatch in node counts is handled via zero-padding. The difference is concatenated with the global feature embeddings and passed through a fully connected network to produce the final binary prediction for \textit{critical} versus \textit{non-critical} changes.

By using shared GNN weights, this Siamese-style architecture allows the model to learn a consistent encoding of ENC graph structure and systematically compare changes between the old and new datasets while evaluating different GNN architectures, depths, and residual configurations.

\section{Experiments}
To evaluate the impact of different GNN architectures on ENC change classification, all experiments are conducted on the same graph dataset with consistent node features, edge structure, and edge types. This controlled setup ensures that observed performance differences arise from architectural variations rather than changes in the input representation.

\textbf{Dataset.} Our dataset consists of 9281 ENC changes. These changes were reviewed by maritime navigation experts and a total of 6069 changes were deemed \textit{critical} while the remaining 3212 were considered \textit{non-critical}. 

\textbf{Evaluation.} We perform five-fold stratified cross-validation on our dataset to evaluate model performance. Each fold contains 80\% of the records for training and 20\% for validation, with splits generated such that the distribution of review labels is preserved across folds. For each fold, the training and validation keys are saved, along with summary statistics describing dataset sizes and label distributions. 

Our primary evaluation metrics are macro F1 score, precision, recall, and overall accuracy. We report the mean and standard deviation of these metrics across the five folds for each architectural configuration, providing a robust estimate of model performance.

\textbf{Implementation Details.} For consistency and reproducibility across all experiments, we explicitly set the random seed. All models are trained for 100 epochs using the Adam optimizer with a fixed learning rate of $1\mathrm{e}{-3}$. To address class imbalance, we use a weighted binary cross-entropy loss, with the positive class weighted according to the ratio of negative to positive samples in the training set.

\begin{table*}[h]
\centering
\caption{Performance of GNN models across depths with and without residual connections.
Values represent the mean $\pm$ standard deviation over 5-fold cross-validation. Bold indicates the best performing variant of each model for the given metric.}
\label{tab:fullresults}
\renewcommand{\arraystretch}{1.15}
\setlength{\tabcolsep}{4pt}

\begin{tabular}{lcccccc}
\toprule
\textbf{Model} &
\textbf{3-Layers} &
\textbf{3-Layers (Res)} &
\textbf{4-Layers} &
\textbf{4-Layers (Res)} &
\textbf{5-Layers} &
\textbf{5-Layers (Res)} \\
\midrule
\multicolumn{7}{c}{\textbf{Accuracy}} \\
\midrule
MLP        & 0.8080 $\pm$ 0.0055 & ---                & ---                & ---                & ---                & ---                \\
GCN        & 0.8193 $\pm$ 0.0045 & 0.8473 $\pm$ 0.0058 & 0.8130 $\pm$ 0.0070 & 0.8453 $\pm$ 0.0048 & 0.8111 $\pm$ 0.0123 & \textbf{0.8545 $\pm$ 0.0057} \\
GAT        & 0.8205 $\pm$ 0.0077 & \textbf{0.8423 $\pm$ 0.0086} & 0.8194 $\pm$ 0.0040 & 0.8320 $\pm$ 0.0051 & 0.8158 $\pm$ 0.0030 & 0.8172 $\pm$ 0.0060 \\
GT         & 0.8319 $\pm$ 0.0051 & 0.8309 $\pm$ 0.0054 & 0.8293 $\pm$ 0.0052 & \textbf{0.8320 $\pm$ 0.0057} & 0.8274 $\pm$ 0.0034 & 0.8288 $\pm$ 0.0042 \\
GraphSAGE  & 0.8745 $\pm$ 0.0051 & 0.8576 $\pm$ 0.0037 & 0.8765 $\pm$ 0.0041 & 0.8757 $\pm$ 0.0050 & \textbf{0.8766 $\pm$ 0.0050} & 0.8713 $\pm$ 0.0062 \\
\midrule

\multicolumn{7}{c}{\textbf{Precision}} \\
\midrule
MLP        & 0.8774 $\pm$ 0.0231 & ---                & ---                & ---                & ---                & ---                \\
GCN        & 0.8031 $\pm$ 0.0055 & 0.8299 $\pm$ 0.0060 & 0.7943 $\pm$ 0.0077 & 0.8281 $\pm$ 0.0047 & 0.7975 $\pm$ 0.0085 & \textbf{0.8372 $\pm$ 0.0059} \\
GAT        & 0.8015 $\pm$ 0.0078 & \textbf{0.8257 $\pm$ 0.0109} & 0.7999 $\pm$ 0.0042 & 0.8144 $\pm$ 0.0056 & 0.7966 $\pm$ 0.0032 & 0.7976 $\pm$ 0.0064 \\
GT         & \textbf{0.8142 $\pm$ 0.0069} & 0.8137 $\pm$ 0.0047 & 0.8106 $\pm$ 0.0057 & 0.8137 $\pm$ 0.0061 & 0.8105 $\pm$ 0.0051 & 0.8100 $\pm$ 0.0043 \\
GraphSAGE  & 0.8630 $\pm$ 0.0082 & 0.8406 $\pm$ 0.0038 & 0.8646 $\pm$ 0.0052 & 0.8631 $\pm$ 0.0061 & \textbf{0.8665 $\pm$ 0.0072} & 0.8571 $\pm$ 0.0064 \\
\midrule

\multicolumn{7}{c}{\textbf{Recall}} \\
\midrule
MLP        & 0.8232 $\pm$ 0.0378 & ---                & ---                & ---                & ---                & ---                \\
GCN        & 0.7996 $\pm$ 0.0047 & 0.8444 $\pm$ 0.0038 & 0.8057 $\pm$ 0.0107 & 0.8476 $\pm$ 0.0060 & 0.8085 $\pm$ 0.0055 & \textbf{0.8486 $\pm$ 0.0080} \\
GAT        & 0.8132 $\pm$ 0.0051 & 0.8290 $\pm$ 0.0062 & 0.8089 $\pm$ 0.0051 & \textbf{0.8350 $\pm$ 0.0085} & 0.8101 $\pm$ 0.0062 & 0.8117 $\pm$ 0.0069 \\
GT         & 0.8228 $\pm$ 0.0049 & \textbf{0.8334 $\pm$ 0.0059} & 0.8225 $\pm$ 0.0048 & 0.8250 $\pm$ 0.0066 & 0.8238 $\pm$ 0.0121 & 0.8244 $\pm$ 0.0063 \\
GraphSAGE  & 0.8600 $\pm$ 0.0032 & 0.8549 $\pm$ 0.0051 & \textbf{0.8623 $\pm$ 0.0060} & 0.8623 $\pm$ 0.0060 & 0.8591 $\pm$ 0.0045 & 0.8593 $\pm$ 0.0078 \\
\midrule

\multicolumn{7}{c}{\textbf{Macro F1}} \\
\midrule
MLP        & 0.8483 $\pm$ 0.0096 & ---                & ---                & ---                & ---                & ---                \\
GCN        & 0.8031 $\pm$ 0.0055 & 0.8354 $\pm$ 0.0050 & 0.7978 $\pm$ 0.0069 & 0.8346 $\pm$ 0.0045 & 0.7975 $\pm$ 0.0085 & \textbf{0.8421 $\pm$ 0.0066} \\
GAT        & 0.8059 $\pm$ 0.0065 & \textbf{0.8269 $\pm$ 0.0079} & 0.8036 $\pm$ 0.0040 & 0.8210 $\pm$ 0.0061 & 0.8014 $\pm$ 0.0029 & 0.8030 $\pm$ 0.0064 \\
GT         & 0.8172 $\pm$ 0.0032 & \textbf{0.8197 $\pm$ 0.0046} & 0.8153 $\pm$ 0.0049 & 0.8180 $\pm$ 0.0055 & 0.8140 $\pm$ 0.0042 & 0.8155 $\pm$ 0.0048 \\
GraphSAGE  & 0.8609 $\pm$ 0.0038 & 0.8461 $\pm$ 0.0036 & \underline{\textbf{0.8632 $\pm$ 0.0046}} & 0.8625 $\pm$ 0.0054 & 0.8624 $\pm$ 0.0050 & 0.8571 $\pm$ 0.0064 \\
\bottomrule

\end{tabular}
\end{table*}

\subsection{Ablations}
To assess the structural stability and depth-scalability of the proposed GNN backbones, we conducted an ablation study varying model depth and connectivity. Specifically, we evaluated four distinct GNN architectures—GCN, GAT, TransformerConv (GT), and GraphSAGE—across depths of 3, 4, and 5 layers. Furthermore, to investigate the impact of over-smoothing and vanishing gradients common in deeper graph networks, we trained each variant both with and without layer-wise residual connections (skip connections). This resulted in a comprehensive evaluation grid designed to isolate the contribution of the graph propagation mechanism versus the depth of the network.

\subsection{Experimental Results}
Table \ref{tab:fullresults} summarizes mean $\pm$ standard deviation for accuracy, precision, recall, and macro-F1 across 5 folds. GraphSAGE achieves the highest accuracy (up to 0.8766 $\pm$ 0.0050) and macro-F1 (up to 0.8632 $\pm$ 0.0046), outperforming the MLP baseline (accuracy 0.8080 $\pm$ 0.0055, macro-F1 0.8483 $\pm$ 0.0096) and all other GNN variants. Performance for GraphSAGE is consistently high and stable across depths. 

Adding layer-wise residual connections yields architecture-dependent effects. For GCN, residual connections produce substantial gains (3-layer macro-F1 0.8031 → 0.8354; 5-layer macro-F1 up to 0.8421), alleviating depth-related degradation. GAT similarly benefits at shallower depths (3-layer macro-F1 0.8059 → 0.8269) but shows mixed results for deeper models. TransformerConv gains are marginal. In contrast, residuals do not improve GraphSAGE and slightly degrade some metrics (3-layer macro-F1 0.8609 → 0.8461), suggesting the skip pathway interferes with GraphSAGE’s native aggregation dynamics.

\section{Discussion}
The experimental results underscore the importance of incorporating graph topology into ENC change criticality classification. The substantial margin by which GraphSAGE outperforms the MLP baseline indicates that change criticality is heavily influenced by the local neighborhood context of the navigational object. The dominance of GraphSAGE suggests that its inductive neighborhood aggregation strategy is particularly well-aligned with this domain. Unlike GCNs, which rely on spectral approximations that can suffer from over-smoothing at greater depths, GraphSAGE’s concatenation-based aggregator appears to preserve distinct node identities more effectively as depth increases. Our ablation study reveals distinct behaviors regarding residual connections. The strong positive impact of residuals on GCN and GAT suggests these models are prone to signal degradation or over-smoothing in this specific graph topology. However, GraphSAGE’s indifference—and occasional aversion—to residual connections implies that its native aggregation mechanism effectively mitigates the vanishing gradient problem in this context, rendering additional skip pathways redundant or even interfering.

\section{Conclusion}
In this work, we targeted the challenge of automating ENC change classifification by proposing a novel graph-based approach. 
We encoded both the old and updated ENC datasets into a graph structure, where spatial objects act as nodes and their semantic and spatial relationships form edges. We then framed this task as a graph-pair classification problem. 
The experimental results demonstrate that incorporating graph structure is valuable for this task. Among the tested architectures, GraphSAGE consistently achieves the highest accuracy and macro-F1, suggesting that its inductive neighborhood aggregation aligns particularly well with ENC spatial semantics. The strong and stable performance across depths indicates that GraphSAGE’s concatenation-based updates preserve node identity and mitigate over-smoothing more effectively than spectral approaches like GCN. These results highlight that incorporating graph structure within a siamese-style GNN framework significantly improves performance, confirming that ENC change criticality depends not just on ENC object class but also on their spatial and relational context. Future work will expand this direction through more extensive benchmarking across alternative graph constructions, node feature choices, and GNN architectures.

\section{Acknowledgements}
We acknowledge that this manuscript has been authored by UT-Battelle, LLC under Contract No. DE-AC05-00OR22725 with the U.S. Department of Energy. The United States Government retains and the publisher, by accepting the article for publication, acknowledges that the United States Government retains a non-exclusive, paid-up, irrevocable, world-wide license to publish or reproduce the published form of this manuscript, or allow others to do so, for United States Government purposes. DOE will provide public access to these results of federally sponsored research in accordance with the DOE Public Access Plan (http://energy.gov/downloads/doe-public-access-plan).

\small
\bibliographystyle{IEEEtranN}
\bibliography{references}

@article{kipf2016semi,
  title={Semi-supervised classification with graph convolutional networks},
  author={Kipf, TN},
  journal={arXiv preprint arXiv:1609.02907},
  year={2016}
}

@article{hamilton2017inductive,
  title={Inductive representation learning on large graphs},
  author={Hamilton, Will and Ying, Zhitao and Leskovec, Jure},
  journal={Advances in neural information processing systems},
  volume={30},
  year={2017}
}

@inproceedings{velivckovic2018graph,
  title={Graph Attention Networks},
  author={Veli{\v{c}}kovi{\'c}, Petar and Cucurull, Guillem and Casanova, Arantxa and Romero, Adriana and Li{\`o}, Pietro and Bengio, Yoshua},
  booktitle={International Conference on Learning Representations},
  year={2018}
}

@article{gorishniy2021revisiting,
  title={Revisiting deep learning models for tabular data},
  author={Gorishniy, Yury and Rubachev, Ivan and Khrulkov, Valentin and Babenko, Artem},
  journal={Advances in neural information processing systems},
  volume={34},
  pages={18932--18943},
  year={2021}
}

@article{shi2020masked,
  title={Masked label prediction: Unified message passing model for semi-supervised classification},
  author={Shi, Yunsheng and Huang, Zhengjie and Feng, Shikun and Zhong, Hui and Wang, Wenjin and Sun, Yu},
  journal={arXiv preprint arXiv:2009.03509},
  year={2020}
}

@article{bromley1993signature,
  title={Signature verification using a" siamese" time delay neural network},
  author={Bromley, Jane and Guyon, Isabelle and LeCun, Yann and S{\"a}ckinger, Eduard and Shah, Roopak},
  journal={Advances in neural information processing systems},
  volume={6},
  year={1993}
}

@misc{NOAA_ENC,
  author       = {{NOAA Office of Coast Survey}},
  title        = {{NOAA ENC – Electronic Navigational Charts}},
  howpublished = {\url{https://www.nauticalcharts.noaa.gov/charts/noaa-enc.html}},
  note         = {Accessed: 2025-11-30},
  year         = {n.d.}
}

@article{scarselli2008graph,
  title={The graph neural network model},
  author={Scarselli, Franco and Gori, Marco and Tsoi, Ah Chung and Hagenbuchner, Markus and Monfardini, Gabriele},
  journal={IEEE transactions on neural networks},
  volume={20},
  number={1},
  pages={61--80},
  year={2008},
  publisher={IEEE}
}

\end{document}